%% file: main.tex
\documentclass[sigconf,screen]{acmart}
\AtBeginDocument{%
  }

\copyrightyear{2025}
\acmYear{2025}
\setcopyright{acmlicensed}\acmConference[RecSys '25]{Proceedings of the Nineteenth ACM Conference on Recommender Systems}{September 22--26, 2025}{Prague, Czech Republic}
\acmBooktitle{Proceedings of the Nineteenth ACM Conference on Recommender Systems (RecSys '25), September 22--26, 2025, Prague, Czech Republic}
\acmDOI{10.1145/3705328.3748047}
\acmISBN{979-8-4007-1364-4/2025/09}

\begin{document}
\title{Off-Policy Evaluation and Learning for Matching Markets}

\author{Yudai Hayashi}
\orcid{0000-0001-7952-8606}
\affiliation{%
  \institution{Wantedly, inc.}
  \city{Tokyo}
  \country{Japan}
}
\email{yudai@wantedly.com}

\author{Shuhei Goda}
\orcid{0009-0005-9656-1420}
\affiliation{%
  \institution{Independent Researcher}
  \city{Tokyo}
  \country{Japan}
}

\email{shuhei.goda@gmail.com}

\author{Yuta Saito}
\orcid{0000-0003-4357-5835}
\affiliation{%
  \institution{Cornell University}
  \city{New York}
  \country{United States}
}
\email{ys552@cornell.edu}

\renewcommand{\shortauthors}{Hayashi et al.}

\begin{abstract}
Matching users based on mutual preferences is a fundamental aspect of services driven by reciprocal recommendations, such as job search and dating applications. Although A/B tests remain the gold standard for evaluating new policies in recommender systems for matching markets, it is costly and impractical for frequent policy updates. Off-Policy Evaluation (OPE) thus plays a crucial role by enabling the evaluation of recommendation policies using only offline logged data naturally collected on the platform. However, unlike conventional recommendation settings, the large scale and bidirectional nature of user interactions in matching platforms introduce variance issues and exacerbate reward sparsity, making standard OPE methods unreliable. To address these challenges and facilitate effective offline evaluation, we propose novel OPE estimators, \textit{DiPS} and \textit{DPR}, specifically designed for matching markets. Our methods combine elements of the Direct Method (DM), Inverse Propensity Score (IPS), and Doubly Robust (DR) estimators while incorporating intermediate labels, such as initial engagement signals, to achieve better bias-variance control in matching markets. Theoretically, we derive the bias and variance of the proposed estimators and demonstrate their advantages over conventional methods. Furthermore, we show that these estimators can be seamlessly extended to offline policy learning methods for improving recommendation policies for making more matches. We empirically evaluate our methods through experiments on both synthetic data and A/B testing logs from a real job-matching platform. The empirical results highlight the superiority of our approach over existing methods in off-policy evaluation and learning tasks for a variety of configurations.

\end{abstract}

\maketitle

\section{Introduction}
Online platforms for job search and dating increasingly influence how people find employment and establish personal connections~\cite{su2022optimizing}. These services rely heavily on the effectiveness of recommendation policies, as user satisfaction and engagement directly depend on accurate matching~\cite{tomita2023fast,tomita2022matching}. To help users efficiently find suitable matches, these platforms should recommend user pairs with mutual preferences. However, achieving accurate and efficient matching presents unique challenges due to the large-scale and bidirectional nature of interactions. To address these challenges, researchers have developed recommendation methods specifically designed for matchmaking~\cite{tomita2023fast,kleinerman2018optimally,palomares2021reciprocal}.
To identify the most effective method for maximizing the number of matches, practitioners often rely on A/B tests. However, A/B tests impose substantial costs in time, resources, and risk of exposing users to suboptimal policies. Therefore, in practice, off-policy evaluation (OPE) plays a crucial role in evaluating recommender systems~\cite{Saito2021Counterfactual}. OPE enables us to estimate the performance of a new policy using only interaction data naturally collected under an existing policy (e.g., the recommender system currently deployed in production)~\cite{Saito2021Counterfactual,uehara2022review}. An accurate OPE pipeline allows us to identify better recommendation policies before conducting A/B tests, thereby improving efficiency of the tests by focusing only on policies already validated as promising~\cite{Dudik2014doubly, Su2020Doubly, Saito2024Long}.

Recent advances in OPE research have led to the development of numerous estimators and policy gradient methods~\cite{Saito2021Counterfactual,uehara2022review}, most of which are based on inverse propensity scoring (IPS), the direct method (DM) using reward regression, or their hybrid approach, Doubly Robust (DR)~\cite{Dudik2014doubly,Su2020Doubly,su2019cab}. More specifically, IPS applies importance weighting to account for the distributional shift between the new and old (logging) policies, enabling an unbiased estimation of policy performance (such as the expected number of matches). However, IPS suffers from extremely high variance, particularly when the number of actions is large or when reward feedback is sparse, both of which are common in matching platforms. DM, on the other hand, does not rely on importance weights but instead uses a reward regression model to estimate the policy performance. As a result, DM avoids the variance issues but often introduces bias depending on the accuracy of the regression model~\cite{jeunen2021pessimistic}. DR combines IPS and DM, leveraging the unbiasedness of IPS and the lower variance of DM. However, since DR still depends on importance weighting, it remains vulnerable to variance issues in cases of large action spaces, policy divergence, and noisy rewards~\cite{Saito2022Offpolicy}.

\begin{figure}
\centering
    \includegraphics[width=.9\linewidth]{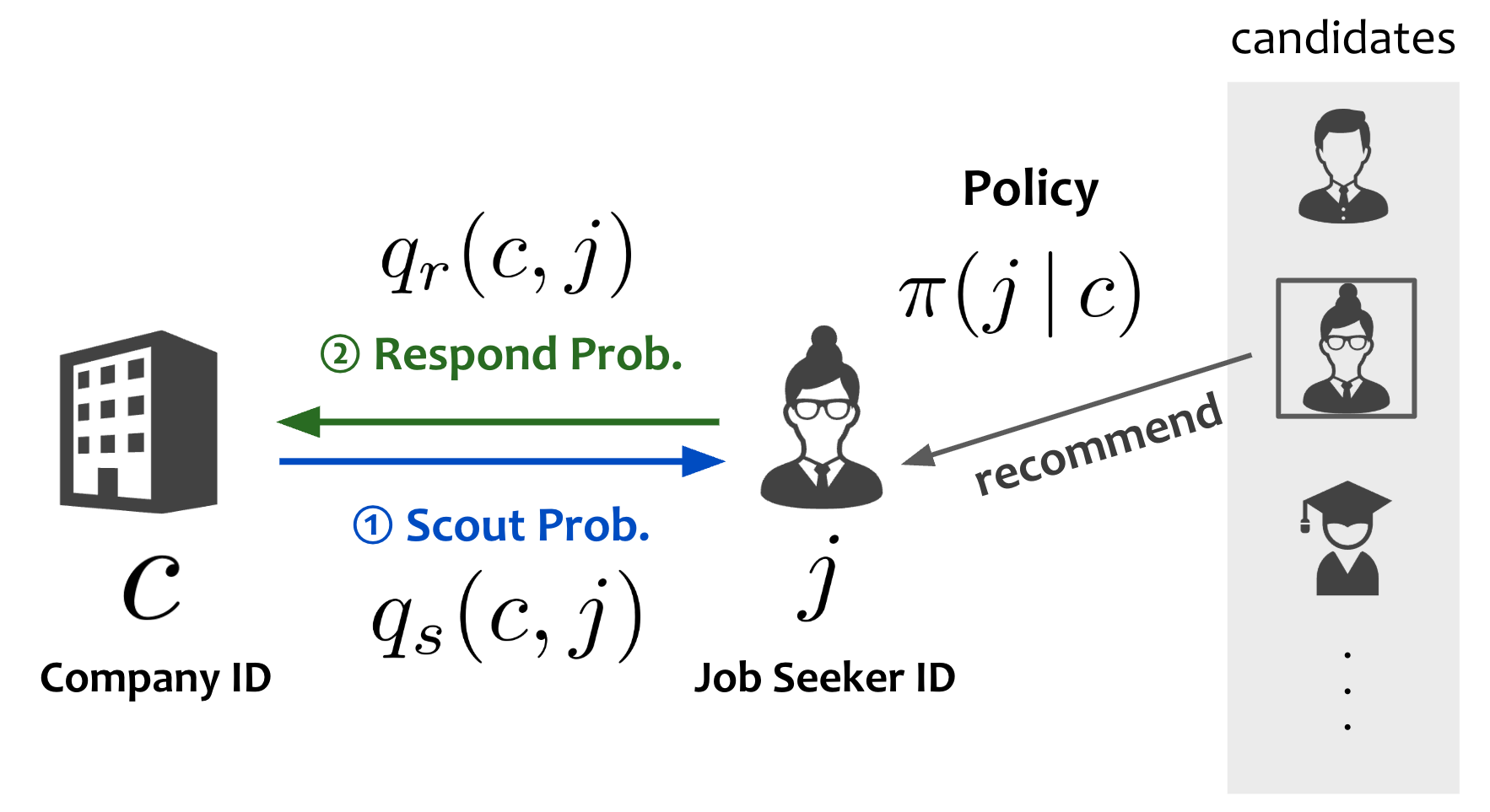}
    \caption{An example situation of our problem of reciprocal recommendation. The platform recommends a job seeker for companies, and then companies decide whether to send a scout to the job seeker. The job seeker who receives a scout then decides whether to respond. A successful match occurs only when a response from the job seeker is observed.}
    \label{fig:problem} \label{-2mm}
\end{figure}

Although these conventional estimation strategies perform well when large amounts of logged data and dense reward observations are available, OPE in matching recommendation settings significantly deviates from these ideal conditions. The most fundamental challenge is that a successful match (i.e., a positive reward) occurs only when both users mutually express interest. Figure~\ref{fig:problem} illustrates an example of the reciprocal recommendation problem. The platform recommends a job seeker for companies, which then decide whether to send scouting messages (\textbf{first-stage reward}). Each job seeker who receives a scout then chooses whether to respond (\textbf{second-stage reward}). A successful match occurs only when a job seeker responds to a scouting message. This leads to an environment with significantly sparse reward signals, making standard OPE methods (such as IPS, DM, and DR) unreliable.

A particular room for improvement in existing estimators in the matching domain is their inability to leverage the first-stage rewards, such as the sending of scouting messages in the example from Figure~\ref{fig:problem}. These methods rely solely on sparse match labels as reward signals, leading to increased variance relative to the expected policy performance and reduced accuracy in policy evaluation and selection. To overcome, we propose two novel OPE estimators, DiPS (\textit{\textbf{Di}rect and \textbf{P}ropensity \textbf{S}core}) and DPR (\textit{\textbf{Di}rect, \textbf{P}ropensity, and doubly \textbf{R}obust}), specifically designed for evaluating recommendation policies in matching platforms. They are the brand new hybrids of the existing estimators where the key idea is the explicit use of the first-stage rewards. Specifically, DiPS applies importance weighting to estimate the first-stage reward under the new policy while imputing the second-stage reward using a regression model, jointly estimating the expected number of matches. DPR builds on and enhances DiPS by further reducing variance, depending on the accuracy of the direct match prediction model.

We conduct a theoretical analysis of our proposed estimators, demonstrating their advantages in terms of bias-variance control under the matching market formulation. Additionally, we show how DiPS and DPR can be extended as policy gradient estimators to implement efficient off-policy learning for matchmaking systems. We finally conduct extensive empirical evaluations using both synthetic data and real-world datasets from the job-matching platform \textit{Wantedly Visit}\footnote{https://www.wantedly.com}. The results demonstrate that our proposed methods enable more accurate policy evaluation, selection, and learning compared to existing approaches by leveraging both the first- and second-stage rewards explicitly. Our real-world experiments using production A/B testing datasets further validate that our methods improve the ability to predict real A/B testing results using only offline logged data, demonstrating their practical significance.

\section{OPE for Matching Markets}
This section formally formulates the problem of OPE for matching markets. In particular, we consider a job-matching platform as an example,\footnote{Note that this does not mean that our formulation is limited to this motivating example regarding the job matching problem. Our formulation can be applied to other matching problems such as real estate and dating platforms.} where companies view job seekers recommended by a policy and send scouting requests when a job seeker aligns with their preferences. On the job seeker's side, they respond to these requests only if they also find the company appealing. A successful interaction, or \textit{\textbf{"match"}}, occurs when the job seeker responds positively to a scouting request, thus tending to be sparse.

Let $j \in \mathcal{J}$ be a job seeker and $c \in \mathcal{C}$ be a company index. A potentially stochastic recommendation policy $\pi (j | c)$ represents the probability of recommending job seeker $j$ to company $c$. We then introduce three types of binary rewards: $s$, $r$, and $m$. The variable $s$ represents the first-stage reward, where $s=1$ if the company sends a scouting request to a job seeker and $s=0$ otherwise. The variable $r$ represents the second-stage reward, where $r=1$ if the job seeker responds to the scouting request. Finally, $m$ represents the ultimate reward, which takes a value of $1$ only when a successful match between a company and a job seeker is observed, meaning $m = s \cdot r$. Given a context vector $x$, we assume that these rewards are drawn from some unknown conditional distributions: $p(s|c,j)$ and $p(m|c,j, s)$. For these reward variables, we define two types of q-functions: $q_{s}(c, j):=\mathbb{E}[s \mid c, j]$ and $q_{r}(c, j)=\mathbb{E}[r \mid c, j, s=1]$.\footnote{For ease of exposition, we consider only company and job seeker indices, but, our formulations can be extended to incorporate their respective features, $x_c$ and $x_j$.} The function $q_{s}(c, j)$ represents the probability that company $c$ sends a scouting request to job seeker $j$. The function $q_{r}(c, j)$ represents the conditional probability that job seeker $j$ responds to company $c$'s scouting request, given that $c$ has sent a scouting request to $j$. By multiplying these two q-functions, we can ultimately define the probability of observing a successful match between $j$ and $c$ as $q_{m}(c, j):=q_{s}(c, j) \cdot q_{r}(c, j)$.

In OPE, our goal is to accurately estimate the policy value, which is defined in our work as follows:
\begin{align} 
    V(\pi) := \frac{1}{|\mathcal{C}|} \sum_{c \in \mathcal{C}} \sum_{j \in \mathcal{J}} \pi (j | c) \cdot q_{s}(c, j) \cdot q_{r}(c, j). \label{eq:Value}
\end{align}
The policy value represents the expected number of matches between job seekers and companies that we obtain under the deployment of a recommendation policy $\pi$ and serves as a meaningful performance measure for policy $\pi$.
    
To implement OPE, we rely on logged bandit data naturally collected under the old or logging policy $\pi_0$ (where $\pi_0 \neq \pi$), denoted as $\mathcal{D}:=\{(j_c,s_c,r_c)\}_{c\in\mathcal{C}}$. The dataset consists of $n$ independent observations drawn from the data distribution induced by the logging policy $\pi_0$. The data generating process is formally given by
\begin{align} 
    p(\mathcal{D}) = \prod_{c \in \mathcal{C}} \pi_0(j_c|c) p(s_c|c,j_c) p(r_c|c,j_c,s_c).  \label{eq:LoggedData}
\end{align}

Multiple metrics exist for evaluating the accuracy of OPE estimators \cite{Kiyohara2024Towards}. In this work, we mainly quantify the accuracy of an estimator $\hat{V}$ using the \textit{mean squared error} (MSE):
\begin{align} 
    \mathrm{MSE}(\hat{V}; \pi) &:= \mathbb{E}_{p(\mathcal{D})}\left[(\hat{V}(\pi; \mathcal{D}) - V(\pi))^2\right], \label{eq:MSE}
\end{align}
which measures the expected squared difference between the estimated value $\hat{V}(\pi; \mathcal{D})$ and the true policy value $V(\pi)$. The MSE can be decomposed into the squared bias and variance as below.
\begin{align} 
    \mathrm{MSE}(\hat{V}; \pi) = \mathrm{Bias}(\hat{V};\pi)^2 + \mathrm{Var}(\hat{V};\pi), 
\end{align} 
where the bias term is defined as $\mathbb{E}[\hat{V}(\pi;\mathcal{D})] - V(\pi)$, and the variance term is given by $\mathbb{E}[(\mathbb{E}[\hat{V}(\pi;\mathcal{D})] - \hat{V}(\pi;\mathcal{D}))^2]$.

While MSE serves as a fundamental measure of estimator accuracy, it does not capture the practical reliability of the estimator in ranking and selecting superior policies. Thus, we additionally consider the \textit{ErrorRate} as a crucial metric, which quantifies how often an estimator misidentifies the relative value of two policies:
\begin{align} 
    &\mathrm{ErrorRate}\left[\hat{V};\pi\right] \label{eq:ErrorRate} \\
    &:= \mathbb{E}_{p(\mathcal{D})}\left[\mathbb{I}\left\{\hat{V}(\pi;\mathcal{D}) \ge \hat{V}(\pi_0;\mathcal{D})\right\}\mathbb{I}\left\{V(\pi) < V(\pi_0)\right\}\right] \notag \\
    &\quad + \mathbb{E}_{p(\mathcal{D})}\left[\mathbb{I}\left\{\hat{V}(\pi;\mathcal{D}) < \hat{V}(\pi_0;\mathcal{D})\right\}\mathbb{I}\left\{V(\pi) \ge V(\pi_0)\right\}\right]. \notag 
\end{align} 
The ErrorRate is defined as the sum of the type-I and type-II error rates, and represents the probability of making a mistake in policy selection. An estimator with a low MSE but a high ErrorRate may still result in suboptimal policy selection. Therefore, it is crucial to design OPE estimators that achieve both low MSE and low ErrorRate to ensure reliable policy evaluation in practice.

\subsection{Typical Estimators}
In the OPE literature, DM, IPS, and DR are widely used as baseline methods. In this section, we analyze the bias and variance of these estimators under the matching market formulation.

\subsubsection{The Direct Method (DM)}
DM estimates the policy value based on the estimation of the match probabilities $q_m(c,j)$ as
\begin{align} 
    \hat{V}_\mathrm{DM}(\pi;\mathcal{D}) 
    = \frac{1}{|\mathcal{C}|} \sum_{c\in\mathcal{C}}\mathbb{E}_{\pi(j|c)}\left[\hat{q}_m(c, j)\right], \label{eq:DM}
\end{align} 
where $\hat{q}_m(c, j)$ is a prediction model for matching probabilities between $c$ and $j$ trained on $\mathcal{D}$. A key characteristic of DM is its low variance. However, it is highly susceptible to bias, which occurs depending on the accuracy of $\hat{q}_m(c, j)$. Particularly in environments with large action spaces and sparse rewards, DM produces high bias due to the difficulty of achieving accurate regression~\cite{saito2022off}.

\subsubsection{Inverse Propensity Scoring (IPS)}
The IPS estimator, unlike DM, does not rely on prediction models. Instead of modeling match probabilities directly, it applies importance weighting to correct for bias. Given logged data collected under $\pi_0$, IPS is defined as:
\begin{align} 
    \hat{V}_\mathrm{IPS}(\pi;\mathcal{D}) 
    = \frac{1}{|\mathcal{C}|}\sum_{c\in\mathcal{C}} \underbrace{\frac{\pi(j_c|c)}{\pi_0(j_c|c)}}_{:=w(c,j)}\cdot m_c, \label{eq:IPS} 
\end{align} 
where $w(c,j):= \pi(j|c)/\pi_0(j|c)$ is called the importance weight, which plays a crucial role in correcting for the distributional shift and ensuring unbiasedness. There exist situations where the logging policy $\pi_0$ is unknown, and in such a circumstance, we need to estimate it by learning a supervised classifier to estimate the probabilities of observing $j$ give $c$ using the logged data $\mathcal{D}$.

Unlike DM, IPS has zero bias (i.e., $\mathbb{E}_{p(\mathcal{D})} [\hat{V}_\mathrm{IPS}(\pi;\mathcal{D})] = V(\pi)$) under the common support condition and known logging policy:
\begin{align*} 
  \text{common support:} \quad  \pi(j|c) > 0 \Rightarrow \pi_0(j|c) > 0, \quad \forall j\in \mathcal{J}, 
\end{align*}
which ensures the sufficient exploration by the logging policy $\pi_0$. 

However, when the logging policy is estimated, IPS produces bias depending on the accuracy of the estimation.\footnote{The bais of IPS under an estimated logging policy $\hat{\pi}_0(j|c)$ is given as 
\begin{align}
    \mathrm{Bias}(\hat{V}_\mathrm{IPS}(\pi;\mathcal{D}) ) = \frac{1}{|\mathcal{C}|} \sum_{c \in \mathcal{C}} \mathbb{E}_{\pi(j|c)} \left[ \left(\frac{\pi_0(j|c)}{\hat{\pi}_0(j|c)} - 1 \right)  q_m(c,j) \right] \label{eq:IPSBiasWithPiHat},
\end{align} which becomes zero when $\pi_0(j|c) = \hat{\pi}_0(j|c)$.} Moreover, a major limitation of IPS is its high variance, particularly when the target policy differs significantly from the logging policy or when the action space is large \cite{Saito2022Offpolicy,saito2023off,kiyohara2024off}. More specifically, we can represent the variance of IPS under our formulation as follows.

\begin{proposition} \label{prop:IPSVariance}
The variance of the IPS estimator under the matching market formulation is represented as follows.
\begin{align}
    \operatorname{Var}\left[\hat{V}_{\mathrm{IPS}}(\pi ; \mathcal{D})\right]
    =\frac{1}{|\mathcal{C}|^{2}} \sum_{c \in \mathcal{C}}\{ & \mathbb{E}_{\pi(j \mid c)}\left[w^{2}(c, j) \cdot \sigma_{m}^{2}(c, j)\right] \label{eq:IPSVariance} \\ 
    & \left.+\mathbb{V}_{\pi(j \mid c)}\left[w(c, j) \cdot q_{m}(c, j)\right]\right\}, \notag
\end{align}
which we can derive by applying the law of total variance. $\sigma_{m}^{2}(c, j) := \operatorname{Var} [m |c,j]$ is the conditional variance, or noise, of the match label.
\end{proposition}

From the variance expression, we can see that it depends on the square and variance of the importance weight $w(c, j)$, which causes the typical variance issue of IPS. This problem is further amplified under sparse reward conditions. In such settings, reward signals become increasingly noisy relative to their expected values.\footnote{The variance of a Bernoulli random variable $m \in \{0,1\}$ with expectation parameter $q$ is $q(1 - q)$. The variance relative to the expectation, i.e., $q(1 - q)/q = 1 - q$, increases as $q$ becomes small, which is characteristic of sparse environments.} Consequently, the first term in the variance, which depends on $\sigma_{m}^{2}(c, j)$, becomes particularly problematic, resulting in highly unstable estimations in OPE for matching markets.

\subsubsection{Doubly Robust (DR)} 
The DR estimator combines DM and IPS to leverage their respective strengths. This approach helps remain unbiased as IPS under common support and known logging policy while controlling variance more effectively than either estimator alone. The form of the estimator is specifically given by:
\begin{align} 
    &\hat{V}_\mathrm{DR}(\pi;\mathcal{D}) \label{eq:DR} \\
    &= \frac{1}{|\mathcal{C}|}\sum_{c\in\mathcal{C}}\left\{\frac{\pi(j_c|c)}{\pi_0(j_c|c)}(m_c - \hat{q}_m(c, j_c)) 
    + \mathbb{E}_{\pi(j|c)}\left[\hat{q}_m(c, j)\right]\right\}. \notag 
\end{align}

This estimator is called "doubly robust" because it remains unbiased if at least one of \(q_m(c, j)\) or the importance weights \(w(c, j)\) is estimated correctly even when the logging policy is unknown. Moreover, it often achieves reduced variance compared to IPS.

\begin{proposition} \label{prop:DRVariance}
The variance of the DR estimator under the matching market formulation is represented as follows.
\begin{align}
    \operatorname{Var}\left[\hat{V}_{\mathrm{DR}}(\pi ; \mathcal{D})\right]
    =\frac{1}{|\mathcal{C}|^{2}} \sum_{c \in \mathcal{C}}\{ & \mathbb{E}_{\pi(j \mid c)}\left[w^{2}(c, j) \cdot \sigma_{m}^{2}(c, j)\right] \label{eq:DRVariance} \\ 
    & \left.+\mathbb{V}_{\pi(j \mid c)}\left[w(c, j) \cdot \Delta_{q_{m}, \hat{q}_{m}}(c, j)\right]\right\}, \notag
\end{align}
where $\Delta_{q_{m}, \hat{q}_{m}}(c, j):= q_{m}(c,j) - \hat{q}_m(c,j)$ is an estimation error of the match probability estimator $\hat{q}_m(c,j)$.
\end{proposition}

We can see from the variance expression of DR that the second term is dependent on the accuracy of the q-function estimator, i.e., $\Delta_{q_{m}, \hat{q}_{m}}(c, j)$, which is expected to be small compared to the original q-function $q_m(c,j)$ unless the match probability estimation $\hat{q}_m(c,j)$ is highly inaccurate. However, as previously discussed, sparse rewards negatively affect the first term in the variance expression, which remains unchanged for DR as well. Sparse rewards also make it difficult to learn an accurate $\hat{q}_m(c,j)$, limiting the variance reduction advantage. Consequently, despite its theoretical benefits against IPS, DR can suffer from high MSE when applied to the evaluation of recommender systems in matching markets.

\section{RELATED WORK}
This section summarizes key related studies.

\paragraph{\textbf{Reciprocal Recommender System}}

A reciprocal recommender system is highly effective in domains where bidirectional preferences play a crucial role, such as matching platforms \cite{Yang2024Revisiting}. In particular, reciprocal recommendation approaches are widely used in dating services \cite{Li2012Meet, Tu2014Online, Pizzato2010Recon, Neve2019Latent, Luo2020RRCN, Xia2015Reciprocal} and job-matching platforms \cite{Chen2023Bilateral, Hu2023Boss, yildirim2021bideepfm, Goda2024Best, Mine2013Reciprocal, su2022optimizing}.

Various architectures for reciprocal recommender systems have been proposed. A well-known approach is training two separate models to predict preferences in each direction and then aggregating their predictions using functions such as the harmonic mean. More recently, deep learning techniques have been employed to capture the complex preferences of both parties more accurately \cite{Yild2021biDeepFM}. Additionally, methods based on Graph Neural Networks \cite{Lai2024Knowledge, Liu2024Linksage, Luo2020RRCN} have been explored to enhance recommendation performance. Some studies have also formulated reciprocal recommendation within the framework of sequential recommendation to better model dynamic user preferences \cite{Zheng2023Reciprocal}.

Our primary focus is not on developing algorithms for reciprocal recommender systems but on formulating and designing accurate methods for off-policy evaluation in matching markets for the first time. By leveraging our methods, practitioners working on reciprocal recommender systems will be able to reliably identify the most suitable algorithms among the many proposed in academia, using only their offline logged interactions.

\paragraph{\textbf{Off-Policy Evaluation and Learning}}
Off-policy evaluation (OPE) and learning (OPL) have gained particular attention in contextual bandit and reinforcement learning settings as they offer a safe and cost-efficient alternative to online A/B tests ~\cite{mehrotra2018towards,gilotte2018offline,saito2021evaluating,Kiyohara2024Towards}.

Numerous OPE methods have been extensively studied \cite{Yi2020Adaptive, Yi2019Cab, Lichtenberg2023Double} to enable accurate evaluation of decision-making policies. However, the reward prediction model in DM and the importance weights in IPS can become unstable under certain conditions such as less data, noisy rewards, and large action spaces, preventing them from being effective in the matching market setup. Recent efforts have reduced MSE by decreasing bias in the DM term \cite{Farajtabar2018More, kang2007demystifying, Thomas2016data}, reducing variance in importance weighting \cite{Bembom2008Data, Wang2017Optimal, Lichtenberg2023Double, Saito2022Offpolicy, Shimizu2024Effective, Metelli2021Subgaussian, Saito2023OffPolicy}, or combining multiple estimators \cite{Wang2017Optimal, Farajtabar2018More}.

A straightforward approach to mitigating the issue of exploding importance weights is called clipping, which restricts the maximum value of the weights to a predefined threshold \cite{Bembom2008Data,Yi2019Cab}. While this method effectively reduces variance coming from the variation of the importance weights, it introduces bias and does not directly deal with the sparsity of the reward. \citet{Saito2022Offpolicy} introduced the marginalized importance weight as a technique to substantially reduce variance compared to conventional IPS leveraging embeddings in the action space. The Switch-DR estimator \cite{Wang2017Optimal} smoothly interpolates between DM and DR via a hyperparameter $\lambda$, allowing for the mitigation of excessive MSE growth by appropriately tuning the parameter. Even though we focus on DM, IPS, and DR as the baseline estimators, Appendix A demonstrates that we can readily extend our methods to improve more sophisticated estimators such as Switch-DR~\cite{Wang2017Optimal} and MIPS~\cite{Saito2022Offpolicy} as well.

When learning new policies offline, these OPE estimators serve as estimators for the policy gradient~\cite{Saito2021Counterfactual}. By performing iterative gradient ascents based on an estimated policy gradient, we can learn effective policies in ideal scenarios, such as when logged data is abundant, rewards are densely observed, and the action space is relatively small~\cite{jeunen2021pessimistic,liang2022local}. However, under those challenging scenarios, estimation of the policy gradient becomes unstable, causing performance lag in OPL as well as OPE.

To the best of our knowledge, no prior research has explicitly addressed the challenges of OPE and OPL in matching markets. This work takes the first step by formulating the problems of OPE and OPL for matching markets and discussing the limitations of directly applying existing methods. We also develop highly effective methods for OPE and OPL that explicitly leverage first-stage rewards that we can observe in matching markets and provide both theoretical and empirical analyses of their effectiveness.

\section{The Proposed APPROACH}
In the previous sections, we observed that typical estimators in OPE do not account for the bidirectional nature and reward sparsity of matching platforms. To address these limitations, we explicitly leverage the useful structure of matching platforms, specifically the existence of first-stage rewards $s$, not just the ultimate rewards $m$.

\subsection{The DiPS Estimator}
We first propose the \textbf{DiPS} estimator, a novel hybrid of the DM and IPS estimators, specifically designed for the matching problem.
More specifically, DiPS estimates the expected first-stage reward $q_s(c,j)$ (e.g., the probability of company $c$ sending a scouting request to job seeker $j$) by applying IPS to the first-stage reward observation $s$, achieving its unbiased estimate. For the second-stage reward, DiPS relies on a reward regression model $\hat{q}_r(c,j)$ trained on offline logged data $\mathcal{D}$, similarly to DM. By employing this novel hybrid approach, DiPS avoids fully applying IPS to the sparse match label $m$, thereby mitigating the variance problem. At the same time, it does not rely entirely on DM, preventing potential bias issues. 

The DiPS estimator is rigorously defined as follows:
\begin{align} 
    \hat{V}_\mathrm{DiPS}(\pi;\mathcal{D}) 
    = \frac{1}{|\mathcal{C}|}\sum_{c\in\mathcal{C}} \left(\frac{\pi(j_c|c)}{\pi_0(j_c|c)} \cdot s_c \right) \cdot \hat{q}_r(c, j_c). \label{eq:DiPS}
\end{align}
where $w(c,j):= \pi(j|c)/\pi_0(j|c)$ is the same importance weight used in IPS and DR.

As in Eq.~\eqref{eq:DiPS}, the DiPS estimator estimates the expected first-stage reward function by applying IPS to the first-stage label $s_c$. It then estimates the expected second-stage reward using the reward model $\hat{q}_r(c, j_c)$. By leveraging IPS and DM to estimate different stages of the rewards, DiPS achieves desirable bias-variance control in matching markets compared to typical estimators.

We first analyze the bias of the DiPS estimator as follows.

\begin{theorem} \label{thm:DiPSBias}
The bias of the DiPS estimator under the matching market formulation is represented by the following equation:
\begin{align} 
    \mathrm{Bias}\left[\hat{V}_\mathrm{DiPS}(\pi;\mathcal{D})\right]  = \frac{1}{|\mathcal{C}|}\sum_{c\in\mathcal{C}}\mathbb{E}_{\pi(j|c)}\left[q_s(c,j)\cdot\Delta_{q_{r}, \hat{q}_{r}}(c, j)\right], \label{eq:DiPSBias}
\end{align}
where $\Delta_{q_{r}, \hat{q}_{r}}(c, j):= \hat{q}_{r}(c,j) - q_r(c,j)$ is an estimation error of the expected second-stage reward estimator $\hat{q}_r(c,j)$.
\end{theorem}
The bias analysis suggests that the bias of DiPS is characterized by the estimation error of the prediction model for the second-stage reward, $\hat{q}_r(c,j)$. This is reasonable because DiPS provides an unbiased estimate of the first-stage reward by applying importance weighting, leaving bias only in the second-stage reward estimation. This bias is expected to be smaller than that of DM since it directly estimates the match probability $q_m(c,j)$, which is highly sparse and difficult to estimate accurately.

Next, the following analyzes the bias of DiPS under an estimated logging policy $\hat{\pi}_0(j|c)$.
\begin{theorem}
\label{thm:DiPSBiasWithPiHat}
The bias of the DiPS estimator with an estimated logging policy $\hat{\pi}_0(j|c)$ is represented by the following equation:
\begin{align} 
    & \mathrm{Bias}\left[\hat{V}_\mathrm{DiPS}(\pi;\mathcal{D},\hat{\pi})\right] \label{eq:DiPSBiasWithPiHat} \\
    & = \frac{1}{|\mathcal{C}|}\sum_{c\in\mathcal{C}}\mathbb{E}_{\pi(j|c)}\left[\left( \frac{\pi_0(j|c)}{\hat{\pi}_0(j|c)} \frac{\hat{q}_r(c,j)}{q_r(c,j)} - 1 \right) q_m(c,j) \right] \notag .
\end{align}
\end{theorem}
Comparing Eq.~\eqref{eq:DiPSBiasWithPiHat} to the bias of IPS in Eq.~\eqref{eq:IPSBiasWithPiHat}, it is interesting that the bias of DiPS can be smaller than that of IPS when the estimation errors against $\pi_0(j|c)$ and $q_r(c,j)$ are positively correlated. 

We then derive the variance of the DiPS estimator.
\begin{theorem} \label{thm:DiPSVariance}
The variance of the DiPS estimator under the matching market formulation is given as follows:
\begin{align} 
    \mathrm{Var}\left[\hat{V}_\mathrm{DiPS}(\pi;\mathcal{D})\right] 
    &= \frac{1}{|\mathcal{C}|^2}\sum_{c\in\mathcal{C}} \Big\{\mathbb{E}_{\pi(j|c)}\left[w^2(c,j)\cdot\sigma_s^2(c,j)\cdot\hat{q}_r^2(c,j)\right] \notag \\
    & + \mathbb{V}_{\pi(j|c)}\left[w(c,j)\cdot q_{s}(c,j)\cdot\hat{q}_{r}(c,j)\right]\Big\}, \label{eq:DiPSVariance}
\end{align}
where $\sigma_{s}^{2}(c, j) := \operatorname{Var} [s |c,j]$ is the conditional variance of the first-stage reward.
\end{theorem}

Analogous to the variance expression of IPS (Eq.~\eqref{eq:IPSVariance}), the first term in the DiPS variance captures the contribution from the importance weights. A key distinction, however, lies in the fact that the variance in DiPS depends on the noise in the first-stage reward, denoted by $\sigma_s^2(c, j)$, scaled by the square of the second-stage reward model, $\hat{q}_r^2(c, j)$. This quantity is generally smaller than the noise in the sparse match label, $\sigma_m^2(c, j)$, which appears in the IPS variance.
\begin{theorem} \label{thm:VariaceReduction}
The DiPS estimator reduces the variance by at least the following amount compared to IPS, if the second-stage reward estimator is not overestimating, i.e., $\hat{q}_r(c,j) \le q_r(c,j), \forall (c,j)$.
\begin{align}
     & \mathrm{Var}\left[\hat{V}_\mathrm{IPS}(\pi;\mathcal{D})\right] - \mathrm{Var}\left[\hat{V}_\mathrm{DiPS}(\pi;\mathcal{D})\right] \notag \\
     & \ge \frac{1}{|\mathcal{C}|^2}\sum_{c\in\mathcal{C}} \mathbb{E}_{\pi(j|c)}\left[w^2(c,j)\cdot q_s(c,j)\cdot \sigma_r^2(c,j)\right]  \ge 0 \label{eq:VariaceReduction}
\end{align}
\end{theorem}
The theorem indicates that DiPS generally achieves lower variance than IPS. It is intriguing that the variance reduction achieved by DiPS becomes particularly large when the square of the importance weight $w^2(c, j)$, the expected first-stage reward $q_s(c, j)$, and the noise in the second-stage reward $\sigma_r^2(c, j)$ are large. This highlights a particularly advantageous property of DiPS in challenging scenarios of large action spaces or significant policy divergence.

\subsection{The DPR Estimator}
The previous section introduced DiPS as a new hybrid approach that explicitly leverages the first-stage reward in the matching market setup. This section further extends DiPS by incorporating a model for the expected match probability, $\hat{q}_m(c,j)$, originally used in DM and DR, to achieve even better bias-variance control. The following formulation realizes this idea, defining the new DPR estimator:
\begin{align}
    & \hat{V}_\mathrm{DPR}(\pi;\mathcal{D}) \label{eq:DPR} \\
    & = \frac{1}{|\mathcal{C}|}\sum_{c\in\mathcal{C}}\left\{\frac{\pi(j_c|c)}{\pi_0(j_c|c)}(s_c\cdot\hat{q}_r(c,j) - \hat{q}_m(c,j)) + \mathbb{E}_{\pi(j|c)}\left[\hat{q}_m(c,j)\right]\right\}, \notag 
\end{align}
which explicitly incorporates the first-stage reward and leverages both the prediction models for the expected second-stage reward and ultimate reward, $\hat{q}_r(c,j)$ and $\hat{q}_m(c,j)$. An interesting interpretation of DPR is that it extends DiPS by the same fundamental idea as DR yet additionally leveraging the first-stage reward $s$.

The following analyzes the key statistical properties of DPR.
\begin{theorem} \label{thm:DPRBias}
The bias of the DPR estimator under the matching market formulation is derived as follows.
\begin{align} 
    & \mathrm{Bias}\left[\hat{V}_\mathrm{DPR}(\pi;\mathcal{D})\right] = \mathrm{Bias}\left[\hat{V}_\mathrm{DiPS}(\pi;\mathcal{D})\right]. \label{eq:DiPSBias}
\end{align}
\end{theorem}
The above theorem suggests that, theoretically, DPR produces exactly the same bias as DiPS. This result is reasonable because DPR incorporates the additional reward model $\hat{q}_m(c,j)$, similar to DR, in a way that does not introduce additional bias.

Next, we analyze the variance of DPR.
\begin{theorem} \label{thm:DPRVariance}
The variance of the DPR estimator under the matching market formulation is derived as follows.
\begin{align} 
    \mathrm{Var}\big[\hat{V}_\mathrm{DPR}  (\pi;\mathcal{D} & )\big]
    = \frac{1}{|\mathcal{C}|^2}\sum_{c\in\mathcal{C}} \Big\{\mathbb{E}_{\pi(j|c)}\left[w^2 (c,j) \cdot\sigma_s^2(c,j) \cdot\hat{q}_r^2(c,j)\right] \notag \\
    & + \mathbb{V}_{\pi(j|c)}\left[w(c,j)\cdot q_s(c,j)\cdot\Delta_{q_r,\hat{q}_r}(c,j)\right] \Big\} \label{eq:DPRVariance},
\end{align}
\end{theorem}
It is interesting that the second term in the variance expression of DPR is characterized by the estimation error of the model $\hat{q}_r(c, j)$, i.e., $\Delta_{q_r,\hat{q}_r}(c,j)$, as opposed to $q_r(c,j)$ in DiPS. Therefore, as long as the prediction model for the second-stage reward is reasonably accurate, DPR produces lower variance than DiPS.

\subsection{Extension to OPL for Matching Markets}
In addition to the OPE counterpart, we can formulate the problem of learning a new policy for matchmaking to optimize the expected reward as  $\max_{\theta} V(\pi_{\theta})$ where $\theta \in \mathbb{R}^{d}$ is the policy parameter. A common approach to solving this policy learning problem is the policy-based method, which updates the policy parameter iteratively using gradient ascent: $\theta_{t+1} \leftarrow \theta_{t} + \eta \cdot \nabla_{\theta}V(\pi_{\theta})$ where $\eta > 0$ is the learning rate. Since the true gradient
\begin{equation} 
    \nabla_{\theta}V(\pi_{\theta}) = \sum_{c\in\mathcal{C}} \mathbb{E}_{\pi_{\theta}(j|c)}[q_m(c,j)\nabla_{\theta}\log\pi_{\theta}(j|c)] \notag
\end{equation}
is unknown, it must be estimated from logged data. This estimation has been addressed in the existing literature by applying DM, IPS, or DR estimators~\cite{su2019cab,Metelli2021Subgaussian}. However, similar to OPE for matching markets, policy gradient estimators based on these conventional methods suffer from either bias or high variance in our problem.

To address these challenges in OPL for matching markets, the following extends DiPS to serve as a policy gradient estimator to optimize the policy value using logged data.
\begin{align}
    \nabla_{\theta}\hat{V}_{\mathrm{DiPS}}(\pi_{\theta};\mathcal{D}) := \frac{1}{|\mathcal{C}|}\sum_{c\in\mathcal{C}} \left(\frac{\pi(j_c|c)}{\pi_0(j_c|c)} \cdot s_c \right) \hat{q}_r(c, j_c) s_{\theta}(c,j) \label{eq:DiPS-PG}
\end{align}
where $s_{\theta}(c,j) := \nabla_{\theta}\log\pi_{\theta}(j|c)$ is the policy score function. Note that the policy gradient estimator based on DPR can similarly be defined, and is described in the appendix.

As discussed earlier, the policy gradient estimator in Eq.~\eqref{eq:DiPS-PG} explicitly incorporates first-stage rewards. By following the same theoretical reasoning as in the OPE setting, our proposed gradient estimators achieve more efficient OPL for matching markets enjoying better bias-variance control in the gradient estimation.

\section{SYNTHETIC DATA EXPERIMENT}

This section empirically evaluates the proposed methods using synthetic datasets for a variety of situations. 

\paragraph{\textbf{Setup}}
To generate synthetic data, we begin with characterizing companies ($c \in \mathcal{C}$) and job seekers ($j \in \mathcal{J}$), each by 10-dimensional context vectors ($x_c$ and $x_j$), sampled from the standard normal distribution. We then synthesize the first- and second-stage expected reward functions using non-linear transformations as follows:
\begin{align} 
    q_{s}(c, j) &= \mathrm{sigmoid}\left[(x_c - x_c^2)\theta_s + (x_c^3 + x_c^2 - x_c)M_s x_j^T - \theta_\mathrm{sp}b_s\right], \notag \\
    q_{r}(c, j) &= \mathrm{sigmoid}\left[(x_j^3 + x_j^2 - x_j)\theta_r + (x_j - x_j^2)M_r x_c^T - \theta_\mathrm{sp}b_r\right], \notag 
\end{align}
where $\mathrm{sigmoid}(x)$ denotes the sigmoid function, defined as $1 / (1 + \exp(-x))$. These reward functions are parameterized by matrices and vectors $M_s, M_r, \theta_s$, and $\theta_r$, all of which are sampled from the standard normal distribution. The experimental parameter $\theta_\mathrm{sp}$ controls the sparsity of rewards, and $b_s$ and $b_r$ are sampled from a uniform distribution over the range $[0, 2]$. Note that the expected match probability between $c$ and $j$ is defined as $q_{m}(c, j) = q_{s}(c, j) \cdot q_{r}(c, j).$

We construct the logging policy $\pi_0$ by applying the softmax function to $q_m(c, j)$ as
\begin{align} 
    \pi_0(j|c) = \frac{\exp(\beta\cdot q_m(c,j))}{\sum_{j'\in J} \exp(\beta\cdot q_m(c,j'))}, 
\end{align}
where $\beta$ is an experimental parameter that controls the optimality and entropy of the logging policy. We set $\beta = -0.5$ by default.

In contrast, we define the target policy $\pi(j|c)$ using an epsilon-greedy strategy:
\begin{align} 
    \pi(j|c) = (1-\epsilon)\cdot\mathbb{I}\left\{j=\underset{j'\in J}{\operatorname{argmax}} \, q_m(c,j')\right\} + \frac{\epsilon}{|J|},\notag 
\end{align}
where $\epsilon\in[0, 1]$ controls the level of randomness in $\pi(j|c)$. We set $\epsilon = 0.2$ by default.

\begin{figure*}
    \centering
    \includegraphics[width=.9\linewidth]{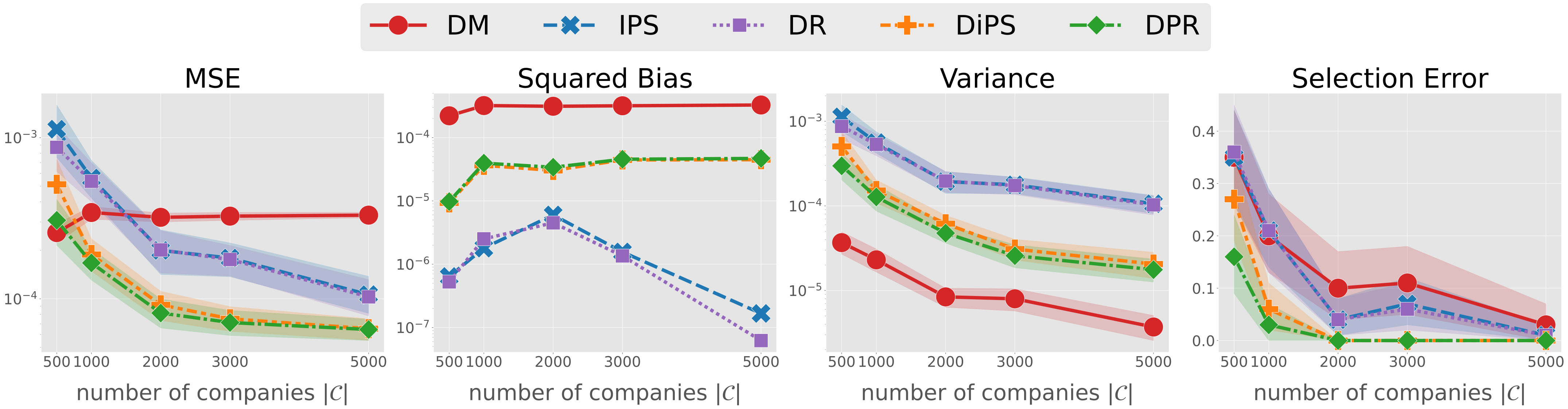} \vspace{-2mm}
     \caption{Performance of estimators when varying the number of companies.} \vspace{-2mm}
    \label{fig:NumCompany}
\end{figure*} 
\begin{figure*}
    \centering
    \includegraphics[width=.9\linewidth]{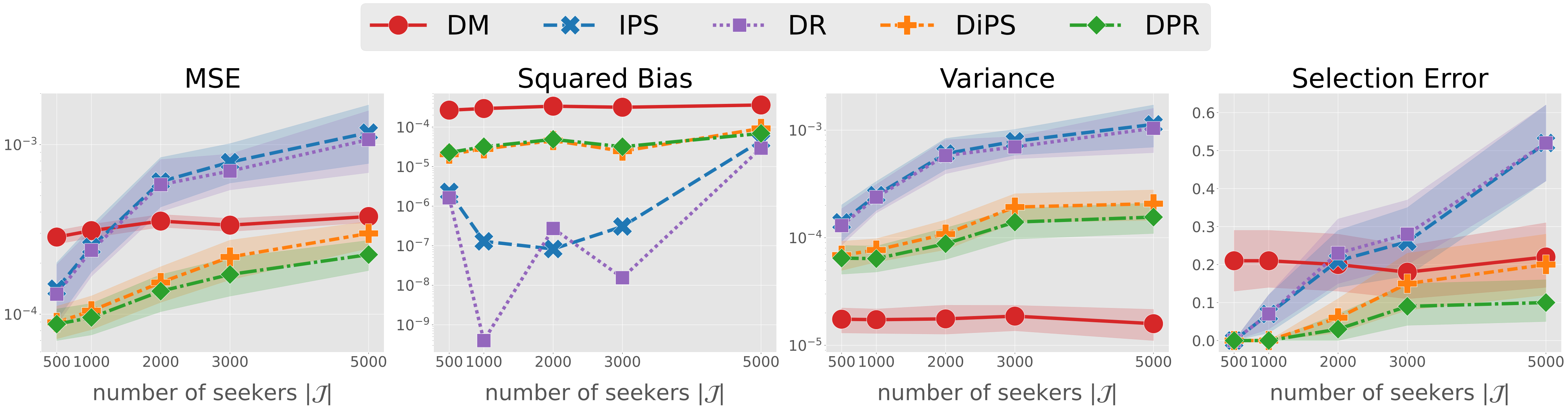} \vspace{-2mm}
    \caption{Performance of estimators when varying the number of job seekers.} \vspace{-2mm}
    \label{fig:NumSeeker}
\end{figure*}
\begin{figure*}
    \centering
    \includegraphics[width=.9\linewidth]{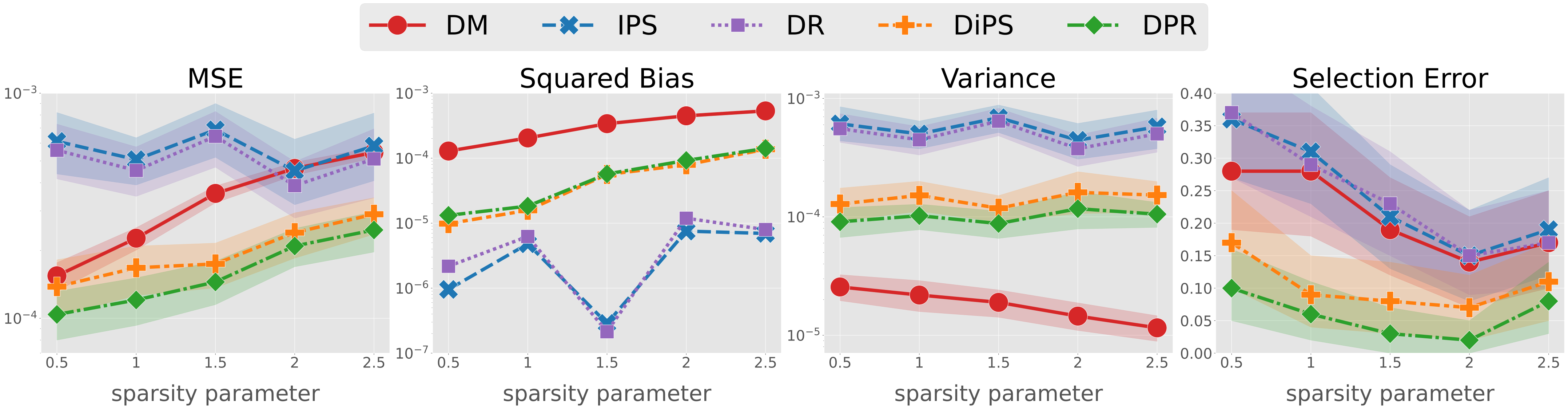} \vspace{-2mm}
    \caption{Performance of estimators when varying reward sparsity. A larger sparsity parameter leads to more severe sparsity.} \vspace{-2mm}
    \label{fig:Sparsity}
\end{figure*}

\paragraph{\textbf{Results}}
In our experiments on the evaluation task, we compared the MSE and ErrorRate of DiPS and DPR against DM, IPS, and DR, which serve as baseline estimators. We also conducted experiments on the learning task, where we compared the policy value of policies learned using DiPS-PG and DPR-PG against DM-PG, IPS-PG, and DR-PG as baselines (PG stands for Policy Gradient). We performed both OPE and OPL experiments by varying the number of companies, the number of job seekers, and reward sparsity parameters. The results are presented in Figures~\ref{fig:NumCompany} to \ref{fig:Sparsity}.

First, we vary the number of companies (i.e., size of the logged data $|\mathcal{D}|$) in Figure~\ref{fig:NumCompany}. The figure shows that every estimator except DM shows a decrease in MSE and variance as the number of companies increases, which is expected. In particular, both DiPS and DPR achieve significantly lower MSE than the baseline methods across most experimental configurations. More specifically, they achieve much lower variance than IPS and DR while introducing slightly higher bias due to their reliance on the estimated second-stage reward function, which aligns with our theoretical analysis. This improvement in MSE for OPE translates to better performance in policy selection tasks (ErrorRate) for DiPS and DPR as well.

Next, we vary the number of job seekers (i.e., size of the action space) in Figure~\ref{fig:NumSeeker}. The results demonstrate the greater robustness of DiPS and DPR in larger action spaces, primarily due to substantial variance reduction compared to IPS and DR. In particular, for larger action spaces, IPS and DR, which heavily depend on importance weighting, produce substantial variance, whereas DiPS and DPR mitigate this issue at the cost of introducing a small amount of bias by estimating the second-stage q-function by a regression model. Moreover, the bias introduced by DiPS and DPR remains significantly smaller than that of DM, consistently leading to lower MSE against it. Similar to the previous experiment, improved estimation accuracy, particularly for larger numbers of job seekers, results in better policy selection performance (lower ErrorRate) for DiPS and DPR compared to the baseline methods.

We also evaluated the performance of the estimators while varying the sparsity of match labels $m$ in Figure~\ref{fig:Sparsity}. As sparsity increases, estimating the match probability function $q_m(c,j)$ becomes more challenging, leading to higher MSE for DM. Since DiPS and DPR do not rely solely on the match prediction model as DM does, they perform significantly better across a range of sparsity configurations. Additionally, DiPS and DPR consistently outperform IPS and DR due to the substantial variance reduction achieved by explicitly leveraging the two-stage reward structure. This also results in a lower ErrorRate in the selection task across various sparsity levels.

Finally, we compare the effectiveness of each estimator in policy learning when used for policy gradient estimation in Figure 7 in the appendix. The figures show the resulting policy values relative to that of the logging policy $\pi_0$ (higher values indicate better performance) for policies learned using each method under varying numbers of companies, numbers of job seekers, and reward sparsity levels. The results suggest that the proposed methods improve upon the logging policy even in challenging scenarios, such as when the number of companies is small, the number of job seekers is large, or reward sparsity is severe. Compared to the baseline methods (DM-, IPS-, and DR-PG), DiPS- and DPR-PG achieve the highest policy values in most cases, demonstrating their effectiveness not only in policy evaluation and selection but also in learning.

\section{REAL-WORLD DATA EXPERIMENTS}
To evaluate the real-world applicability of DiPS and DPR, we conduct experiments about OPE using real-world A/B testing data collected from \textit{Wantedly Visit}, an industrial job-matching platform with over 4 million registered users and more than 40,000 companies. On this platform, job seekers can apply for jobs, and conversely, recruiters can send scouting requests to job seekers. We use A/B testing data from a recommender system in the Wantedly Visit's scouting service, which recommends job seekers to companies. This system is designed as a reciprocal recommender system and recalculates rankings once per day. The dataset includes records of recommendations, which companies sent scouts to which job seekers, and whether the job seekers responded. \vspace{-2mm}

\paragraph{\textbf{Dataset}}
We processed the online testing data to align the original ranked interaction data with our formulation. Specifically, we regard $c$ as a mix of (recommended date, interaction rank, company ID), and $j$ as a mix of (recommended date, job seeker ID). These definitions include the "recommended date" because the contexts of companies, job seekers, and policies vary from day to day. Additionally, we introduce the notion of "interaction rank," which denotes the position in the ranking at which the company interacted with the job seeker. This framing supports a bandit setting and allows us to distinguish between situations in which a company encounters a job seeker at rank $k$ versus rank $k'$ on the same day (where $k \neq k'$). For this evaluation, we used only data where $k \leq 3$ from the original logged dataset. Table 1 in the appendix shows the statistics of the preprocessed dataset. In our dataset, the first-stage reward $s$ corresponds to the scout request, while the ultimate reward $m$ corresponds to a successful match, that is, the company sends a scout to the job seeker, and the job seeker replies positively.

\begin{figure*}
    \centering
    \includegraphics[width=.9\linewidth]{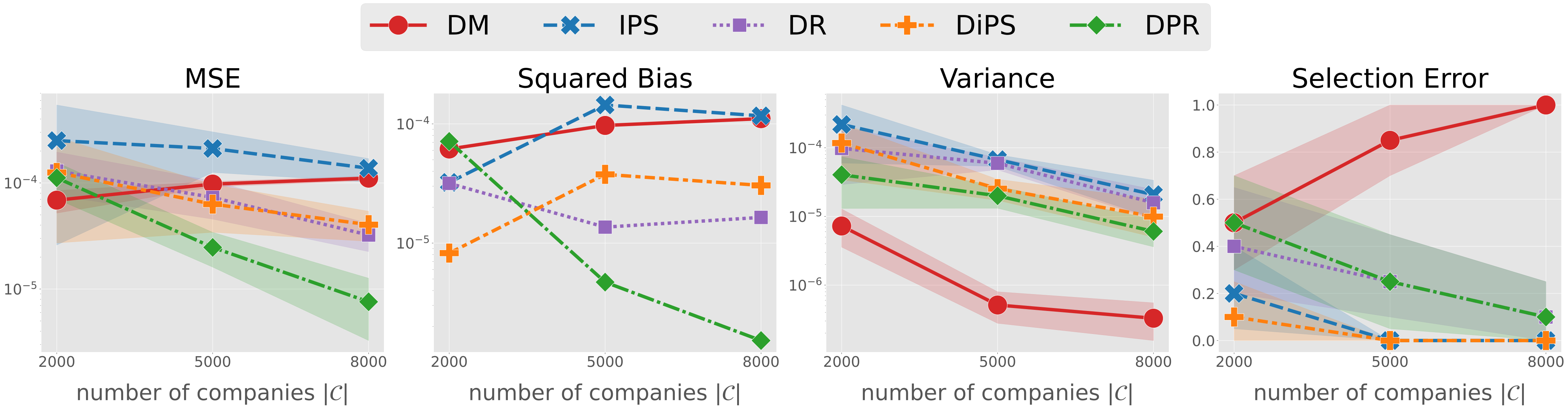} \vspace{-2mm}
    \caption{Performance of estimators when varying the number of companies in read-world data experiment.} \vspace{-2mm}
    \label{fig:RealWorld}
\end{figure*}
\begin{figure}
    \centering
    \includegraphics[width=.675\linewidth]{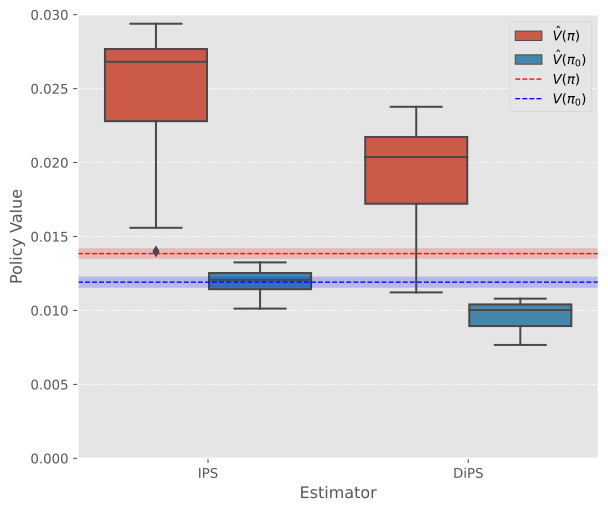} \vspace{-2mm}
    \caption{Distributions of policy values estimated by IPS and DiPS when sampling 8,000 companies.} \vspace{-5mm}
    \label{fig:BoxPlot}
\end{figure}

\paragraph{\textbf{Setup}}
Since we were not able to replicate the exact action choice probabilities under our recommendation policies, we estimated both the logging $\pi_0(j|c)$ and the new $\pi(j|c)$ policies using Gradient Boosting Decision Tree (GBDT) \cite{Jerome2001Greedy}. During the policy estimation task, we treated instances where a job seeker was recommended as a rank-$k$ candidate for a company whose interaction rank was also $k$ as positive examples, and all other instances as negative examples. We trained the models using a 5-fold cross-validation procedure, and then applied the softmax function to normalize the model outputs, ensuring that $\sum_{j\in\mathcal{J}} \pi'(j|c) = 1$ for $\pi'\in\{\pi_0, \pi\}$.

\paragraph{\textbf{Results}}
In our experiments, we compared the MSE and ErrorRate of DiPS and DPR against baseline estimators. During evaluation, we sampled a specific number of companies from the logged data, trained reward estimation models using GBDT within a cross-validation framework, and then estimated the policy value with each method. We repeated this evaluation process 20 times, varying the number of companies in $\mathcal{D}$. Figure~\ref{fig:RealWorld} presents the evaluation results under different company sampling sizes, and Figure~\ref{fig:BoxPlot} shows the distributions of policy values estimated by IPS and DiPS over 20 repetitions with 8,000 companies. We discuss the results below.

First, Figure~\ref{fig:RealWorld} shows that DiPS and DPR achieve low MSE, in particular, DPR achieves significantly lower MSE than the baseline methods across most configurations. This reduction in MSE results from both low bias and low variance. The low variance aligns with our theoretical analysis and findings from synthetic data experiments. In contrast, the bias behavior of DiPS and DPR has two noteworthy characteristics. First, DiPS produces lower bias than IPS, and DPR shows lower bias than DR. As shown in Theorem~\ref{thm:DiPSBiasWithPiHat}, the bias of DiPS can be smaller than that of IPS when the estimation errors of $\hat{\pi}_0(j|c)$ and $\hat{q}_r(c, j)$ are positively correlated. The overestimation observed in $\hat{V}_\mathrm{IPS}(\pi)$ in Figure~\ref{fig:BoxPlot} suggests an underestimation of the logging policy probability $\hat{\pi}_0(j|c)$. Additionally, we observed that the model $\hat{q}_r(c, j)$ tends to underestimate $q_r(c, j)$ during cross-validation. These factors contribute to bias reduction of DiPS. A similar reasoning applies to DPR compared to DR.

The second notable observation is that DPR tends to produce smaller bias than DiPS. Under a known logging policy, both estimators have the same bias, as confirmed by theoretical analysis and synthetic experiments. However, under an unknown logging policy, Eq.~(26) in the appendix shows that the bias of DPR differs from that of DiPS by the term $(1 - \pi_0(j|c) / \hat{\pi}_0(j|c))\hat{q}_m(j|c)$, and in our experiments, the underestimation of the logging policy probability mitigates the bias in DPR. Taken together, these results demonstrate that our proposed methods provide more accurate estimations on real-world data than the baselines by effectively reducing variance and, with unknown logging policies, our methods even reduce bias.

Next, DiPS and DPR also achieve low ErrorRate. An interesting observation is that DiPS and IPS show lower ErrorRate than DPR and DR, despite DPR and DR achieving smaller MSE. Figure~\ref{fig:BoxPlot} helps explain this phenomenon. The overestimation of $V(\pi)$ by IPS contributes to its large MSE but leads to a low ErrorRate, as it more often ranks the target policy above the logging policy, which is correct. DiPS significantly reduces overestimation by lowering both bias and variance, leading to a slight improvement in ErrorRate. Overall, these findings demonstrate that our estimators are well-suited for real-world applications, achieving both low MSE and ErrorRate compared to their respective counterparts.

\section{CONCLUSION}
This paper studies and formulates the problems of off-policy evaluation (OPE) and off-policy learning (OPL) for matching markets for the first time in the relevant literature. Existing estimators, such as DM, IPS, and DR, perform poorly in matching markets due to the large-scale environment and sparse reward nature. We introduce two novel estimators, DiPS and DPR, for off-policy evaluation in the matching domain. These estimators explicitly leverage first-stage reward labels, such as the "sending scout" label in job-matching recommender systems, to achieve better bias-variance control. Extensive experiments on synthetic data and the real-world A/B testing dataset demonstrated that our proposed methods outperformed conventional estimators in policy evaluation, selection, and learning tasks, showing practical impact.

\balance
\bibliographystyle{ACM-Reference-Format}
\bibliography{main.bbl}

\input{appendix}

\end{document}

%% file: appendix.tex
\clearpage
\appendix
\onecolumn

\section{Redesigning Other OPE Estimators for Matching Markets} \label{app:redesign}

Even though, in the main text, we focus on DM, IPS, and DR as the baseline estimators, the following shows that we can readily extend our methods to improve more sophisticated estimators such as Switch-DR~\cite{Wang2017Optimal} and MIPS~\cite{Saito2022Offpolicy}.

First, we rigorously define the Switch-DR~\cite{Wang2017Optimal} estimator under our formulation as follows.
\begin{align}
    \hat{V}_{\mathrm{Switch-DR}}(\pi_{\theta};\mathcal{D},\lambda) 
     := \frac{1}{|\mathcal{C}|}\sum_{c\in\mathcal{C}}\left\{\frac{\pi(j_c|c)}{\pi_0(j_c|c)}(m_c - \hat{q}_m(c, j_c)) \mathbb{I}\{\frac{\pi(j_c|c)}{\pi_0(j_c|c)} \le \lambda\}
    + \mathbb{E}_{\pi(j|c)}\left[\hat{q}_m(c, j)\right]\right\} \label{eq:Switch-DR}.
\end{align}
where $\lambda > 0$ is a hyperparameter. For samples with $\frac{\pi(j_c|c)}{\pi_0(j_c|c)} \le \lambda$, Switch-DR is equal to DR, while it applies DM when $\frac{\pi(j_c|c)}{\pi_0(j_c|c)} > \lambda$.

We can readily extend the Switch-DR estimator to explicitly use the first-stage reward observations $s$ as in our proposed methods.
\begin{align}
    \hat{V}_{\mathrm{Extented Switch-DR}}(\pi_{\theta};\mathcal{D},\lambda) 
     = \frac{1}{|\mathcal{C}|}\sum_{c\in\mathcal{C}}\left\{\frac{\pi(j_c|c)}{\pi_0(j_c|c)}(s_c \cdot \hat{q}_r(c, j_c) - \hat{q}_m(c, j_c)) \mathbb{I}\{\frac{\pi(j_c|c)}{\pi_0(j_c|c)} \le \lambda\}
    + \mathbb{E}_{\pi(j|c)}\left[\hat{q}_m(c, j)\right]\right\} \label{eq:ExtendedSwitch-DR}.
\end{align}

Next, the MIPS estimator~\cite{saito2022off} is known as a method to leverage action embeddings to reduce variance in OPE, particularly for large action spaces. It is defined in our setting as follows.
\begin{align}
    \hat{V}_{\mathrm{MIPS}}(\pi_{\theta};\mathcal{D},\lambda) 
     := \frac{1}{|\mathcal{C}|}\sum_{c\in\mathcal{C}} \frac{\pi(e_{j_c}|c)}{\pi_0(e_{j_c}|c)} \cdot m_c  \label{eq:MIPS}.
\end{align}
where $e_j$ is a embedding of job seeker $j$, and $\pi(e_j|c)$ is a marginalized distribution of job seeker embeddings conditional on $c$.

It is again immediate to extend this MIPS estimator in our matching market formulation to incorporate the two-stage reward structure as follows.
\begin{align}
    \hat{V}_{\mathrm{ExtendedMIPS}}(\pi_{\theta};\mathcal{D},\lambda) 
     := \frac{1}{|\mathcal{C}|}\sum_{c\in\mathcal{C}} \frac{\pi(e_{j_c}|c)}{\pi_0(e_{j_c}|c)} s_c \cdot \hat{q}_r(c, j_c) \label{eq:ExtendedMIPS}.
\end{align}
By applying the same theoretical analysis presented in the main text, these estimators achieve better bias-variance tradeoff in our problem of matching markets due to the use of the two-stage reward structure.

\section{The Policy Gradient Estimator based on DPR}
The following extends the DPR estimator as a policy gradient estimator.
\begin{align}
    \nabla_{\theta}\hat{V}_{\mathrm{DPR-PG}}(\pi_{\theta};\mathcal{D}) 
    = \frac{1}{|\mathcal{C}|}\sum_{c\in\mathcal{C}}\bigg\{\frac{\pi(j_c|c)}{\pi_0(j_c|c)}(s_c\cdot\hat{q}_r(c,j) - \hat{q}_m(c,j)) s_{\theta}(c,j) \notag + \mathbb{E}_{\pi(j|c)}\left[\hat{q}_m(c,j) s_{\theta}(c,j)\right]\bigg\} 
\end{align}

\section{Proofs}

This section provides derivations of the theorems provided in the main text.
\subsection{Propositions~\ref{prop:IPSVariance} and~\ref{prop:DRVariance}}
We first derive the variance of DR via applying the law of total variance below.
\begin{align*}
    \mathbb{V}_{p(\mathcal{D})} \left[ \hat{V}_{D R}(\pi ; \mathcal{D}) \right]
    & = \mathbb{V}_{p(\mathcal{D})} \left[ \frac{1}{|C|} \sum_{c \in C}\left\{\frac{\pi\left(j_{c} \mid c\right)}{\pi_{0}\left(j_{c} \mid c\right)}\left(m_{c}-\hat{q}_{m}\left(c, j_{c}\right)\right)+\mathbb{E}_{\pi(j \mid c)}\left[\hat{q}_{m}(c, j)\right]\right\}\right] \\
    & = \frac{1}{|C|^2} \sum_{c \in C} \mathbb{V}_{\pi_0(j|c)p(m|c,j)} \left[ \left\{\frac{\pi\left(j \mid c\right)}{\pi_{0}\left(j \mid c\right)}\left(m_{c}-\hat{q}_{m}\left(c, j\right)\right)+\mathbb{E}_{\pi(j \mid c)}\left[\hat{q}_{m}(c, j)\right]\right\}\right] \\
    & = \frac{1}{|C|^2} \sum_{c \in C} \mathbb{E}_{\pi_0(j|c)} \left[ \mathbb{V}_{p(m|c,j)} \left[ \left\{\frac{\pi\left(j \mid c\right)}{\pi_{0}\left(j \mid c\right)}\left(m_{c}-\hat{q}_{m}\left(c, j\right)\right)+\mathbb{E}_{\pi(j \mid c)}\left[\hat{q}_{m}(c, j)\right]\right\}\right] \right] \\
    & \quad  + \mathbb{V}_{\pi_0(j|c)} \left[ \mathbb{E}_{p(m|c,j)} \left[ \left\{\frac{\pi\left(j \mid c\right)}{\pi_{0}\left(j \mid c\right)}\left(m_{c}-\hat{q}_{m}\left(c, j\right)\right)+\mathbb{E}_{\pi(j \mid c)}\left[\hat{q}_{m}(c, j)\right]\right\}\right] \right] \\
    & = \frac{1}{|\mathcal{C}|^{2}} \sum_{c \in \mathcal{C}}\{ \mathbb{E}_{\pi(j \mid c)}\left[w^{2}(c, j) \cdot \sigma_{m}^{2}(c, j)\right] +\mathbb{V}_{\pi(j \mid c)}\left[w(c, j) \cdot \Delta_{q_{m}, \hat{q}_{m}}(c, j)\right]\}
\end{align*}
By setting $\hat{q}_m(c,j) = 0$ in the variance expression of DR, we can derive the variance of IPS in Eq.~\eqref{eq:IPSVariance}.

\subsection{Theorems~\ref{thm:DiPSBias} and~\ref{thm:DPRBias}}
We first derive the expectation of DPR below.
\begin{align*}
    \mathbb{E}_{p(\mathcal{D})} \left[ \hat{V}_{DPR}(\pi ; \mathcal{D}) \right]
    & = \mathbb{E}_{p(\mathcal{D})} \left[ \frac{1}{|\mathcal{C}|}\sum_{c\in\mathcal{C}}\left\{\frac{\pi(j_c|c)}{\pi_0(j_c|c)}(s_c\cdot\hat{q}_r(c,j) - \hat{q}_m(c,j)) + \mathbb{E}_{\pi(j|c)}\left[\hat{q}_m(c,j)\right]\right\} \right] \\
    &=  \frac{1}{|\mathcal{C}|}\sum_{c\in\mathcal{C}} \mathbb{E}_{\pi_0(j|c)p(s,r|c,j)} \left[ \left\{\frac{\pi(j_c|c)}{\pi_0(j_c|c)}(s_c\cdot\hat{q}_r(c,j) - \hat{q}_m(c,j)) + \mathbb{E}_{\pi(j|c)}\left[\hat{q}_m(c,j)\right]\right\} \right] \\
    &= \frac{1}{|\mathcal{C}|}\sum_{c\in\mathcal{C}} \mathbb{E}_{\pi_0(j|c)} \left[ \left\{\frac{\pi(j_c|c)}{\pi_0(j_c|c)}(q_s(c,j)\cdot\hat{q}_r(c,j) - \hat{q}_m(c,j)) + \mathbb{E}_{\pi(j|c)}\left[\hat{q}_m(c,j)\right]\right\} \right] \\
    &= \frac{1}{|\mathcal{C}|}\sum_{c\in\mathcal{C}}  \left[ \mathbb{E}_{\pi(j|c)}[ q_s(c,j)\cdot\hat{q}_r(c,j)] - \mathbb{E}_{\pi(j|c)}[\hat{q}_m(c,j))] + \mathbb{E}_{\pi(j|c)}\left[\hat{q}_m(c,j)\right] \right] \\
    &= \frac{1}{|\mathcal{C}|}\sum_{c\in\mathcal{C}} \mathbb{E}_{\pi(j|c)}[ q_s(c,j)\cdot\hat{q}_r(c,j)]
\end{align*}
Therefore, the bias of DPR is
\begin{align*}
    \operatorname{Bias} \left[ \hat{V}_{DPR}(\pi ; \mathcal{D}) \right]
    & = \mathbb{E}_{p(\mathcal{D})} \left[ \hat{V}_{DPR}(\pi ; \mathcal{D}) \right] - V(\pi) \\
    & = \frac{1}{|\mathcal{C}|}\sum_{c\in\mathcal{C}} \mathbb{E}_{\pi(j|c)}[ q_s(c,j)\cdot\hat{q}_r(c,j)] - 
    \frac{1}{|\mathcal{C}|}\sum_{c\in\mathcal{C}} \mathbb{E}_{\pi(j|c)}[ q_s(c,j)\cdot q_r(c,j)] \\
    & = \frac{1}{|\mathcal{C}|}\sum_{c\in\mathcal{C}}\mathbb{E}_{\pi(j|c)}\left[q_s(c,j)\cdot\Delta_{q_{r}, \hat{q}_{r}}(c, j)\right]
\end{align*}
By setting $\hat{q}_m(c,j) = 0$ in the bias expression of DR, we can derive the bias of DiPS in Eq.~\eqref{eq:DiPSBias}.

\subsection{Theorems~\ref{thm:DiPSVariance} and~\ref{thm:DPRVariance}}
We first derive the variance of DR via applying the law of total variance below.
\begin{align*}
    \mathbb{V}_{p(\mathcal{D})} \left[ \hat{V}_{D R}(\pi ; \mathcal{D}) \right]
    & = \mathbb{V}_{p(\mathcal{D})} \left[ \frac{1}{|\mathcal{C}|}\sum_{c\in\mathcal{C}}\left\{\frac{\pi(j_c|c)}{\pi_0(j_c|c)}(s_c\cdot\hat{q}_r(c,j) - \hat{q}_m(c,j)) + \mathbb{E}_{\pi(j|c)}\left[\hat{q}_m(c,j)\right]\right\}\right] \\
    & = \frac{1}{|\mathcal{C}|^2}\sum_{c\in\mathcal{C}} \mathbb{V}_{\pi_0(j|c)p(s,r|c,j)} \left[ \left\{\frac{\pi(j_c|c)}{\pi_0(j_c|c)}(s_c\cdot\hat{q}_r(c,j) - \hat{q}_m(c,j)) + \mathbb{E}_{\pi(j|c)}\left[\hat{q}_m(c,j)\right]\right\}\right] \\
    & = \frac{1}{|\mathcal{C}|^2}\sum_{c\in\mathcal{C}} \mathbb{E}_{\pi_0(j|c)} \left[ \mathbb{V}_{p(s|c,j)} \left[ \left\{\frac{\pi(j_c|c)}{\pi_0(j_c|c)}(s_c\cdot\hat{q}_r(c,j) - \hat{q}_m(c,j)) + \mathbb{E}_{\pi(j|c)}\left[\hat{q}_m(c,j)\right]\right\}\right] \right] \\
    & \quad + \mathbb{V}_{\pi_0(j|c)} \left[\mathbb{E}_{p(s|c,j)} \left[ \left\{\frac{\pi(j_c|c)}{\pi_0(j_c|c)}(s_c\cdot\hat{q}_r(c,j) - \hat{q}_m(c,j)) + \mathbb{E}_{\pi(j|c)}\left[\hat{q}_m(c,j)\right]\right\}\right] \right] \\
    & = \frac{1}{|\mathcal{C}|^2}\sum_{c\in\mathcal{C}}\{\mathbb{E}_{\pi(j|c)}\left[w^2 (c,j) \cdot\sigma_s^2(c,j) \cdot\hat{q}_r^2(c,j)\right] + \mathbb{V}_{\pi(j|c)}\left[w(c,j)\cdot q_s(c,j)\cdot\Delta_{q_r,\hat{q}_r}(c,j)\right]\} 
\end{align*}
By setting $\hat{q}_m(c,j) = 0$ in the variance expression of DPR, we can derive the variance of DiPS in Eq.~\eqref{eq:DiPSVariance}.

\subsection{Theorems~\ref{thm:DiPSBiasWithPiHat} and~\ref{thm:DPRBiasWithPiZeroHat}}

\begin{theorem} \label{thm:DPRBiasWithPiZeroHat}
The bias of the DPR estimator with an estimated logging policy $\hat{\pi}_0(j|c)$ is derived as follows.
\begin{align} 
    \mathrm{Bias}\left[\hat{V}_\mathrm{DPR}(\pi;\mathcal{D})\right] = \frac{1}{|\mathcal{C}|}\sum_{c\in\mathcal{C}}\mathbb{E}_{\pi(j|c)}\left[
    \left( \frac{\pi_0(j|c)}{\hat{\pi}_0(j|c)} - 1 \right) \left( \frac{\hat{q}_r(c,j)}{q_r(c,j)} - \frac{\hat{q}_m(c,j)}{q_m(c,j)} \right) q_m(c,j) +q_s(c,j)\cdot(\hat{q}_r(c,j) - q_r(c,j))
    \right]. \label{eq:DPRBiasWithPiZeroHat}
\end{align}
\end{theorem}

To derive the above theorem, we first derive the expectation of DPR with an estimated logging policy.
\begin{align*}
    & \mathbb{E}_{p(\mathcal{D})} \left[ \frac{1}{|\mathcal{C}|}\sum_{c\in\mathcal{C}}\left\{\frac{\pi(j_c|c)}{\hat{\pi}_0(j_c|c)}(s_c\cdot\hat{q}_r(c,j) - \hat{q}_m(c,j)) + \mathbb{E}_{\pi(j|c)}\left[\hat{q}_m(c,j)\right]\right\} \right] \\
    &= \frac{1}{|\mathcal{C}|}\sum_{c\in\mathcal{C}} \mathbb{E}_{\pi_0(j|c)p(s|c,j)} \left[ \left\{\frac{\pi(j_c|c)}{\hat{\pi}_0(j_c|c)}(s_c\cdot\hat{q}_r(c,j) - \hat{q}_m(c,j)) + \mathbb{E}_{\pi(j|c)}\left[\hat{q}_m(c,j)\right]\right\} \right] \\
    &= \frac{1}{|\mathcal{C}|}\sum_{c\in\mathcal{C}} \mathbb{E}_{\pi(j|c)} \left[ \left\{\frac{\pi_0(j_c|c)}{\hat{\pi}_0(j_c|c)}(q_s(c,j)\cdot\hat{q}_r(c,j) - \hat{q}_m(c,j)) + \mathbb{E}_{\pi(j|c)}\left[\hat{q}_m(c,j)\right]\right\} \right] \\
    &= \frac{1}{|\mathcal{C}|}\sum_{c\in\mathcal{C}} \left\{ \mathbb{E}_{\pi(j|c)} \left[ \frac{\pi_0(j_c|c)}{\hat{\pi}_0(j_c|c)}q_s(c,j)\cdot\hat{q}_r(c,j) \right]  - \mathbb{E}_{\pi(j|c)}\left[ \frac{\pi_0(j_c|c)}{\hat{\pi}_0(j_c|c)} \hat{q}_m(c,j) \right] + \mathbb{E}_{\pi(j|c)}\left[\hat{q}_m(c,j)\right] \right\} \\
\end{align*}
Therefore, the bias of DPR with an estimated logging policy is
\begin{align*}
    & \operatorname{Bias} \left[ \hat{V}_{DPR}(\pi ; \mathcal{D}) \right]
     = \mathbb{E}_{p(\mathcal{D})} \left[ \hat{V}_{DPR}(\pi ; \mathcal{D}) \right] - V(\pi) \\
    &= \frac{1}{|\mathcal{C}|}\sum_{c\in\mathcal{C}} \left\{ \mathbb{E}_{\pi(j|c)} \left[ \frac{\pi_0(j_c|c)}{\hat{\pi}_0(j_c|c)}q_s(c,j)\cdot\hat{q}_r(c,j) \right]  - \mathbb{E}_{\pi(j|c)}\left[ \frac{\pi_0(j_c|c)}{\hat{\pi}_0(j_c|c)} \hat{q}_m(c,j) \right] + \mathbb{E}_{\pi(j|c)}\left[\hat{q}_m(c,j)\right] \right\}
    - 
    \frac{1}{|\mathcal{C}|}\sum_{c\in\mathcal{C}} \mathbb{E}_{\pi(j|c)}[ q_m(c,j)] \\
    & = \frac{1}{|\mathcal{C}|}\sum_{c\in\mathcal{C}} \left\{ \mathbb{E}_{\pi(j|c)} \left[ \frac{\pi_0(j_c|c)}{\hat{\pi}_0(j_c|c)}q_s(c,j)\cdot\hat{q}_r(c,j) - \frac{\pi_0(j_c|c)}{\hat{\pi}_0(j_c|c)} \hat{q}_m(c,j) +\hat{q}_m(c,j)- q_s(c,j) \cdot q_r(c,j) + q_s(c,j) \cdot \hat{q}_r(c,j) - q_s(c,j) \cdot \hat{q}_r(c,j)\right] \right\} \\
    & = \frac{1}{|\mathcal{C}|}\sum_{c\in\mathcal{C}}  \mathbb{E}_{\pi(j|c)} \left[ \left(\frac{\pi_0(j_c|c)}{\hat{\pi}_0(j_c|c)} - 1 \right) \left( q_s(c,j) \cdot \hat{q}_r(c,j) - \hat{q}_m(c,j) \right) + q_s(c,j)\cdot(\hat{q}_r(c,j) - q_r(c,j))\right]  \\
    & = \frac{1}{|\mathcal{C}|}\sum_{c\in\mathcal{C}}\mathbb{E}_{\pi(j|c)}\left[
    \left( \frac{\pi_0(j|c)}{\hat{\pi}_0(j|c)} - 1 \right) \left( \frac{\hat{q}_r(c,j)}{q_r(c,j)} - \frac{\hat{q}_m(c,j)}{q_m(c,j)} \right) q_m(c,j) +q_s(c,j)\cdot(\hat{q}_r(c,j) - q_r(c,j))
    \right]
\end{align*}
By setting $\hat{q}_m(c,j) = 0$ in the bias expression of DPR with an estimated logging policy, we can derive the bias of DiPS with an estimated logging policy in Eq.~\eqref{eq:DiPSBiasWithPiHat}.

\subsection{Theorem~\ref{thm:VariaceReduction}}
The following calculates the difference in the variance of IPS and DiPS using the condition, $\hat{q}_r(c,j) \le q_r(c,j), \forall (c,j)$.
\begin{align*}
    & \mathrm{Var}\left[\hat{V}_\mathrm{IPS}(\pi;\mathcal{D})\right] - \mathrm{Var}\left[\hat{V}_\mathrm{DiPS}(\pi;\mathcal{D})\right] \\
    & = \frac{1}{|\mathcal{C}|^2}\sum_{c\in\mathcal{C}}
    \big\{\mathbb{E}_{\pi(j \mid c)}\left[w^{2}(c, j) \cdot \sigma_{m}^{2}(c, j)\right] - \mathbb{E}_{\pi(j|c)}\left[w^2(c,j)\cdot\sigma_s^2(c,j)\cdot\hat{q}_r^2(c,j)\right] + \mathbb{V}_{\pi(j \mid c)}\left[w(c, j) \cdot q_{m}(c, j)\right] - \mathbb{V}_{\pi(j|c)}\left[w(c,j)\cdot q_{s}(c,j)\cdot\hat{q}_{r}(c,j)\right]\big\} \\
    & \ge \frac{1}{|\mathcal{C}|^2}\sum_{c\in\mathcal{C}}
    \big\{\mathbb{E}_{\pi(j \mid c)}\left[w^{2}(c, j) \cdot \sigma_{m}^{2}(c, j)\right] - \mathbb{E}_{\pi(j|c)}\left[w^2(c,j)\cdot\sigma_s^2(c,j)\cdot q_r^2(c,j)\right] + \mathbb{V}_{\pi(j \mid c)}\left[w(c, j) \cdot q_{m}(c, j)\right] - \mathbb{V}_{\pi(j|c)}\left[w(c,j)\cdot q_{s}(c,j)\cdot q_{r}(c,j)\right]\big\} \\
    & = \frac{1}{|\mathcal{C}|^2}\sum_{c\in\mathcal{C}}
    \big\{\mathbb{E}_{\pi(j \mid c)}\left[w^{2}(c, j) \cdot q_s(c,j) q_r(c,j) (1 - q_s(c,j) q_r(c,j)) \right] - \mathbb{E}_{\pi(j|c)}\left[w^2(c,j)\cdot q_s(c,j)  (1 - q_s(c,j) ) q_r^2(c,j)\right] \big\} \\
    & = \frac{1}{|\mathcal{C}|^2}\sum_{c\in\mathcal{C}} \mathbb{E}_{\pi(j|c)}\left[w^2(c,j)\cdot q_s(c,j)\cdot \sigma_r^2(c,j)\right]
\end{align*}

\begin{figure*}
    \centering
    \includegraphics[width=.8\linewidth]{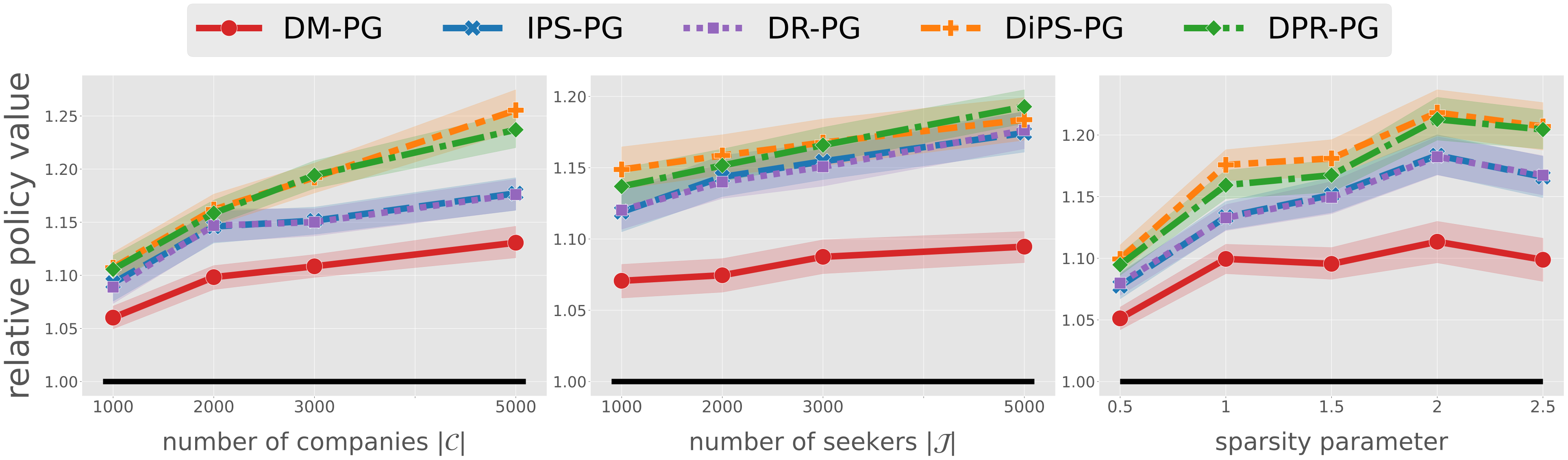}
    \caption{Performance of estimators when used as estimators for policy gradient in OPL.}
    \label{fig:OPL}
\end{figure*}

\label{sub:Data}
\begin{table}[]
    \centering
    \caption{Statistics of the read-world dataset.} \vspace{-3mm}
    \begin{tabular}{cccc}
        \toprule
        $|\mathcal{C}|$&$|\mathcal{J}|$&Reward Sparsity\\
        \midrule
        21,736&17,460&1.2 \% \\
        \bottomrule
    \end{tabular}
    \label{tab:dataset}
\end{table}

\section{Additional Experimental Setups and Results}

\subsection{Performance of Reward Estimation Models}

Table~\ref{tab:RewardModelPerformance} shows the ROC-AUC of our reward prediction models for reply and match. 

\begin{table}[h]
    \centering
    \caption{ROC-AUC of reply and match prediction models when varying the number of sampling companies.}
    \label{tab:RewardModelPerformance} \vspace{-2mm}
    \begin{tabular}{ccc}
        \toprule
        Number of Companies & Reply Prediction Model & Match Prediction Model \\
        \midrule
        2,000 & 0.703 & 0.598\\
        5,000 & 0.680 & 0.693\\
        8,000 & 0.700 & 0.707\\
        \bottomrule
    \end{tabular}
\end{table}

In addition, the following table represents the underestimation in each reward model. The value is $\mathbb{E}[\hat{q}_r(c,j)] / \mathbb{E}[q_r(c, j)] - 1$ and $\mathbb{E}[\hat{q}_m(c, j)] / \mathbb{E}[q_m(c, j)] - 1$ calculated with the logged data.

\begin{table}[h]
    \centering
    \caption{The underestimation of reply and match prediction models when varying the number of sampling companies.}
    \label{tab:RewardModelPerformance} \label{-2mm}
    \begin{tabular}{ccc}
        \toprule
        Number of Companies & Reply Model Underestimation & Match Model Underestimation \\
        \midrule
        2,000 & -23.3 \% & -38.5 \%\\
        5,000 & -34.3 \% & -53.1 \%\\
        8,000 & -38.5 \% & -53.4 \%\\
        \bottomrule
    \end{tabular}
\end{table}

\subsection{Estimated policy value distribution for all estimators}

\begin{figure*}[h]
    \centering
    \includegraphics[width=.6\linewidth]{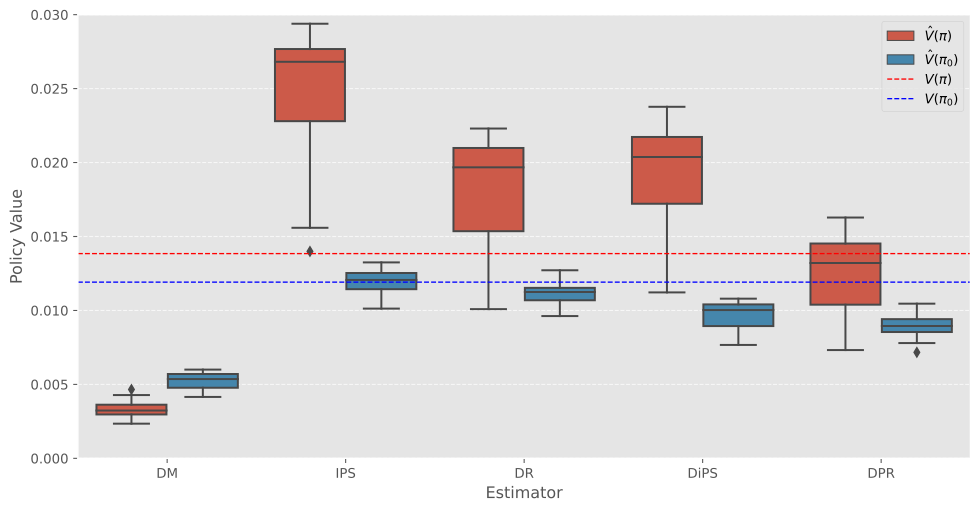}
    \caption{Estimated policy value distribution for all estimators when sampling 8,000 companies.} \vspace{-2mm}
    \label{fig:BoxPlotAll}
\end{figure*}

Figure~\ref{fig:BoxPlotAll} shows the estimated policy value distribution for DiPS, DPR and other baseline estimators. The red and blue box represent the estimated policy distributions and the dashed lines represent true policy values.

